\documentclass[sn-mathphys]{sn-jnl}% Math and Physical Sciences Reference Style
\jyear{2022}%

\theoremstyle{thmstyleone}%
\theoremstyle{thmstyletwo}%

\theoremstyle{thmstylethree}%
\usepackage{soul}
\usepackage{url}
\usepackage[utf8]{inputenc}
\usepackage[small]{caption}
\usepackage{graphicx}
\usepackage{amsmath}
\usepackage{amssymb}
\usepackage{booktabs}

\usepackage{xcolor}

\DeclareMathOperator*{\argmax}{arg\!max}
\DeclareMathOperator*{\argmin}{arg\!min}

\begin{document}

\title[Synthesizing counterfactual policies for recourse with program synthesis]{Synthesizing explainable counterfactual policies for algorithmic recourse with program synthesis}

%%=============================================================%%
%% Prefix	-> \pfx{Dr}
%% GivenName	-> \fnm{Joergen W.}
%% Particle	-> \spfx{van der} -> surname prefix
%% FamilyName	-> \sur{Ploeg}
%% Suffix	-> \sfx{IV}
%% NatureName	-> \tanm{Poet Laureate} -> Title after name
%% Degrees	-> \dgr{MSc, PhD}
%% \author*[1,2]{\pfx{Dr} \fnm{Joergen W.} \spfx{van der} \sur{Ploeg} \sfx{IV} \tanm{Poet Laureate} 
%%                 \dgr{MSc, PhD}}\email{iauthor@gmail.com}
%%=============================================================%%

\author*[1,2]{\fnm{Giovanni} \sur{De Toni}}\email{giovanni.detoni@unitn.it}

\author[1]{\fnm{Bruno} \sur{Lepri}}\email{lepri@fbk.eu}

\author[2]{\fnm{Andrea} \sur{Passerini}}\email{andrea.passerini@unitn.it}

\affil*[1]{\orgname{Fondazione Bruno Kessler}, \orgaddress{\street{Via Sommarive 18}, \city{Trento}, \postcode{38123}, \state{Italy}}}

\affil[2]{\orgdiv{Department of Information Engineering and Computer Science}, \orgname{University of Trento}, \orgaddress{\street{Via Sommarive 9}, \city{Trento}, \postcode{38123}, \country{Italy}}}

\abstract{
Being able to provide counterfactual interventions – sequences of actions we would have had to take for a desirable outcome to happen – is essential to explain how to change an unfavourable decision by a black-box machine learning model (e.g., being denied a loan request). Existing solutions have mainly focused on generating feasible interventions without providing explanations on their rationale. Moreover, they need to solve a separate optimization problem for each user. In this paper, we take a different approach and learn a program that outputs a sequence of explainable counterfactual actions given a user description and a causal graph. We leverage program synthesis techniques, reinforcement learning coupled with Monte Carlo Tree Search for efficient exploration, and rule learning to extract explanations for each recommended action. An experimental evaluation on synthetic and real-world datasets shows how our approach generates effective interventions by making orders of magnitude fewer queries to the black-box classifier with respect to existing solutions, with the additional benefit of complementing them with interpretable explanations. }


%%\pacs[JEL Classification]{D8, H51}

%%\pacs[MSC Classification]{35A01, 65L10, 65L12, 65L20, 65L70}

\maketitle

\section{Introduction}

Counterfactual explanations are very powerful tools to explain the
decision process of machine learning models
\cite{wachter2017counterfactual,karimi2020survey}. They give us the
intuition of what could have happened if the state of the world was
different (e.g., if you had taken the umbrella, you would not have
gotten soaked). Researchers have developed many methods that can
generate counterfactual explanations given a trained model
\cite{wachter2017counterfactual,dandl2020multi,mothilal2020explaining,karimi2020model,guidotti2018lore,stepin2021survey}. However, these methods do not provide any actionable
information about which steps are required to obtain the given
counterfactual. Thus, most of these methods do not enable algorithmic
recourse. Algorithmic recourse describes the ability to provide
``explanations and recommendations to individuals who are unfavourably
treated by automated decision-making systems"
\cite{karimi2021algorithmic}. For instance, algorithmic recourse can
answer questions such as: what actions does a user have to perform to
be granted a loan? Recently, providing feasible algorithmic recourse
has also become a legal necessity \cite{voigt2017GDPR}. Some research
works address this problem by developing ways to generate
counterfactual interventions \cite{karimi2021algorithmic}, i.e.,
sequences of actions that, if followed, can overturn a decision made
by a machine learning model, thus guaranteeing recourse. While being
quite successful, these methods have several limitations. First, they
are purely optimization methods that must be rerun from scratch
for each new user. As a consequence, this requirement prevents their use for real-time interventions' generation. Second, they are expensive in terms of queries to
the black-box classifier and computing time. Last but not least, they fail to explain their recommendations (e.g., why does the model
suggest getting a better degree rather than changing job?). On the
contrary, explainability has been pointed out as a major requirement
for methods generating counterfactual
interventions~\cite{barocas2020hidden}.
 
\begin{figure*}[t]
    \centering
    \includegraphics[width=\textwidth]{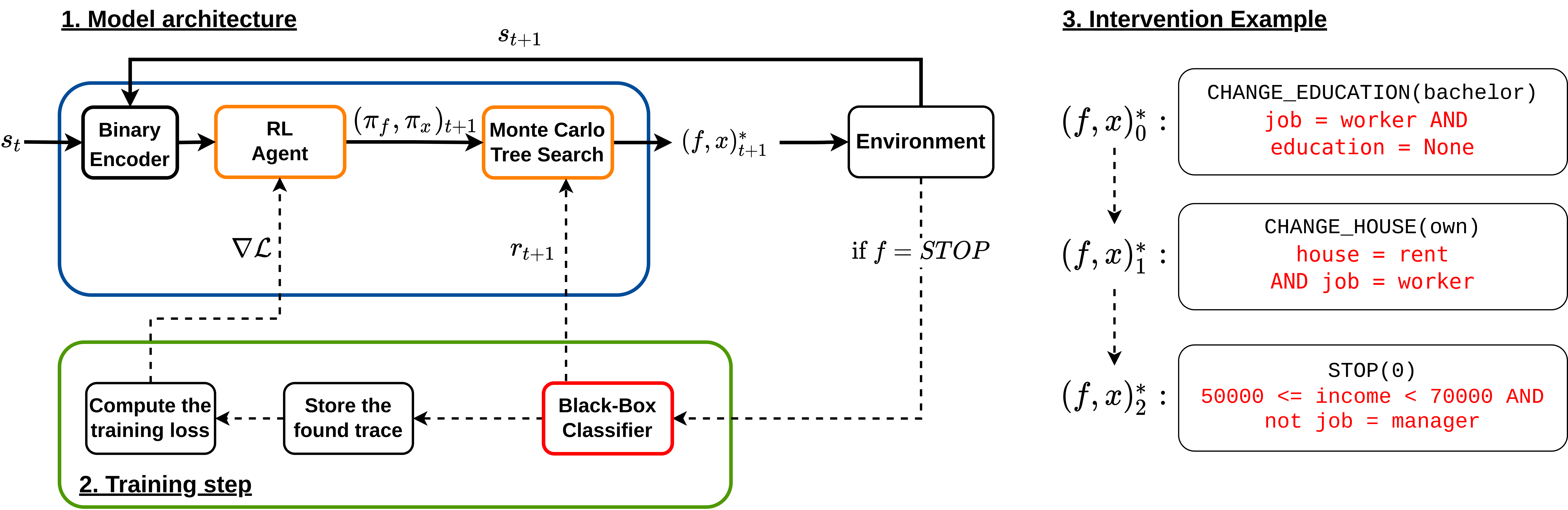}
    \caption{\textbf{1. Model architecture.} Given the state $s_t$ representing the features of the user, the agent generates candidate intervention policies $\pi_{f}$ and $\pi_{x}$ for functions and arguments, respectively (an action is a function-argument pair). 
      MCTS uses these policies as a prior, and it extracts the best next action $(f,
      x)_{t+1}^*$. Once found, the reward received upon making the action is used to improve the MCTS estimates, and 
      correct traces (i.e., those leading to the desired outcome change) are saved in a replay buffer. \textbf{2. Training step.} The buffer is used to sample a subset of correct traces to be used to train the RL agent to mimic the behaviour of MCTS. \textbf{3. Explainable intervention.} Example of an explainable intervention generated by the automaton extracted from the learned agent. Actions are in black, while explanations for each action are in red.
    \label{fig:complete_architecture}}
\end{figure*}

In this paper, we cast the problem of providing explainable
counterfactual interventions as a program synthesis task \cite{detoni2021learning,pierrot2019learning,bunel2018leveraging,BalGauBroNowTar17}: we want to generate a ``program" that provides all
the steps needed to overturn a bad decision made by a machine learning
model. We propose a novel reinforcement learning (RL) method coupled with
a discrete search procedure, Monte Carlo Tree Search
\cite{Coulom06efficientselectivity}, to generate counterfactual
interventions in an efficient data-driven manner. As done by \cite{naumann_consequence-aware_2021}, we assume a causal model encoding relationships between user features and consequences of potential interventions. 
We also provide a solution to distil an explainable deterministic program from the learned policy in the form of an automaton. Fig.~\ref{fig:complete_architecture} provides an overview of the architecture and the learning strategy, and an example of an explainable intervention generated by the extracted automaton. Our approach addresses the
three main limitations that characterize existing solutions:

\begin{itemize}
\item It learns a general policy that can be used to generate interventions
  for multiple users, rather than running separate user-specific optimizations. 

\item By coupling reinforcement learning with Monte Carlo Tree Search,
  it can efficiently explore the search space, requiring massively
  fewer queries to the black-box classifier than the best evolutionary algorithm (EA) model
  available, especially in settings with many features and
  (relatively) long interventions.

\item By extracting a program from the learned policy, it can complement the intervention with explanations
  motivating each action from contextual information. Furthermore, 
  the program can be executed in real-time without accessing the black-box classifier.
\end{itemize}

Our experimental results on synthetic and real-world datasets confirm
the advantages of the proposed solution over existing alternatives in
terms of generality, scalability and interpretability.

\section{Related Work}
Counterfactual explanations are versatile techniques to provide post-hoc interpretability of black-box machine learning models \cite{wachter2017counterfactual,dandl2020multi,mothilal2020explaining,karimi2020model,guidotti2018lore,stepin2021survey}. They are model-agnostic, which means that they can be applied to trained models without performance loss. 
Compared to other global methods \cite{greenwell2018simple, apley2020visualizing}, they provide instead \textit{local} explanations. Namely, they underline only the \textit{relevant factors} impacting a decision for a given initial target instance. They are also human-friendly and present many characteristics of what it is considered to be a \textit{good explanation} \cite{miller2019explanation}. Therefore, they are suitable candidates to provide explanations to end-users since they are both highly-informative and localized. 
Recent research has shown how to generate counterfactual interventions for algorithmic recourse via various techniques \cite{karimi2020survey}, such as probabilistic models \cite{KarKugSchVal20}, integer programming \cite{ustun2019actionable}, reinforcement learning \cite{shavit2019extracting}, program synthesis \cite{ramakrishnan2020synthesizing} and genetic algorithms \cite{naumann_consequence-aware_2021}. Methods with (approximated) convergence guarantees on the optimal counterfactual policies have also been proposed~\cite{tsirtsis2021decisions}. 
However, most of these methods ignore the causal relationships between user features~\cite{tsirtsis2021decisions,ustun2019actionable,shavit2019extracting,ramakrishnan2020synthesizing}.  Without assuming an underlying causal graph, the proposed interventions become \textit{permutation invariant}. For example, given an intervention consisting of three actions $[A,B,C]$,
any intervention that is a permutation of the actions will have the same total cost. More importantly, it has been recently shown that optimal algorithmic recourse is impossible to achieve without a causal model of the interactions between the features~\cite{KarKugSchVal20}.
%\GDT{For example, the work by Ramakrishnan et al.~\cite{ramakrishnan2020synthesizing} also uses program synthesis inspired techniques to provide interventions. However, since they do not have any causality assumption, their models solves a slightly simpler problem, making them unfit to be a potential baseline.}
The work by Karimi et al.~\cite{KarKugSchVal20} does provide algorithmic recourse following a causal model but optimizes for interventions that can operate on multiple user's attributes simultaneously, which is unrealistic. 
CSCF \cite{naumann_consequence-aware_2021} is the only model-agnostic method capable of producing {\it consequence-aware}
sequential interventions by exploiting causal relationships between features represented by a causal graph. However, CSCF is still purely an (evolutionary-based) optimization method, so it has to be run from scratch for each new user. Furthermore, the approach is opaque with respect to the reasons behind a suggested intervention. In this work, we show how our approach improves over CSCF in terms of generality, efficiency and interpretability.

\section{Methods}
\subsection{Problem setting}

\begin{figure}[t!]
    \centering
    \includegraphics[width=\linewidth]{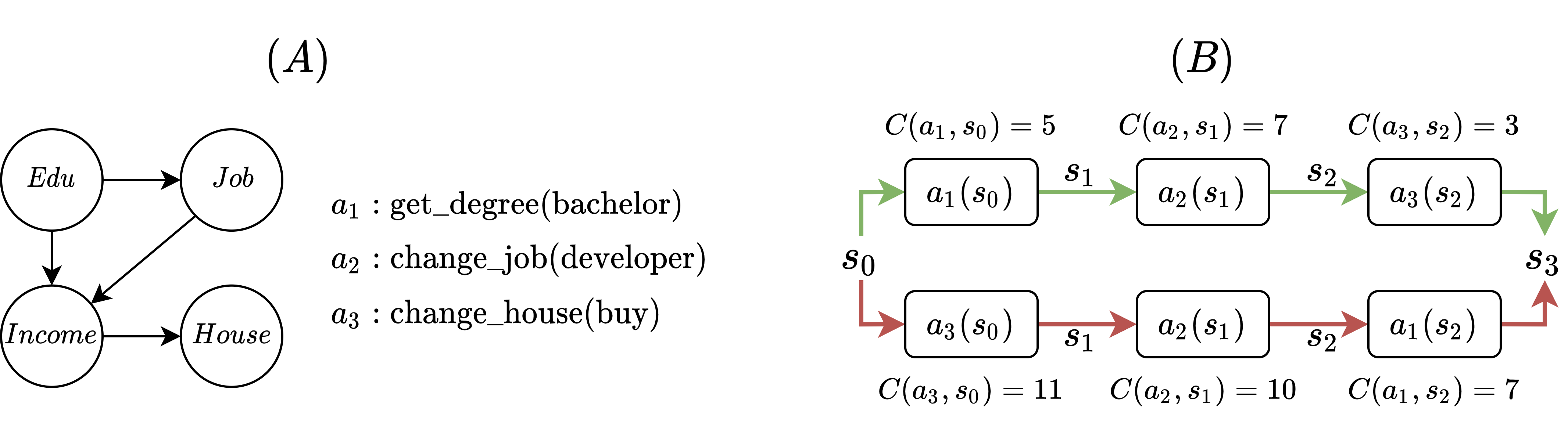}
    \caption{\textbf{Examples of interventions on a causal graph.} (A) A causal graph and a set of candidate actions. (B) Examples of interventions together with their costs. Note that the green line ($\sum C =15$) has a lower cost than the red line ($\sum C=28$) thanks to a better ordering of the actions making up the intervention.}
    \label{fig:causal_graph}
\end{figure}

The state of a user is represented as a vector of attributes
$s\in\mathcal{S}$ (e.g., age, sex, monthly income, job). A
black-box classifier $h : \mathcal{S} \to \{True,False\}$ predicts an
outcome given a user state, with $True$ being favourable to the user
and $False$ being unfavourable. The setting can be easily extended to
multiclass classification by either grouping outcomes in favourable
and unfavourable ones or learning separate programs converting from
one class to the other. A \textit{counterfactual intervention} $I$ is
a sequence of actions. Each action is represented as a tuple,
$(f,x) \in \mathcal{A}$, composed by a \textit{function}, $f$, and its
argument, $x \in \mathcal{X}_f$ (e.g., (\texttt{change\_income},
500)). When an action is performed for a certain user, it modifies their
state by altering one of their attributes according to its argument. A
library $\mathcal{F}$ contains all the possible functions which can be
called. This library and the corresponding DSL (Domain Specific Language) are typically defined as a-priori by experts to prevent changes to protected attributes (e.g., age, sex, etc.). Examples of such DSLs can be found in the Appendix \ref{app:actions}. Moreover, each
function possesses \textit{pre-conditions} in the form of Boolean
predicates over its arguments which describe the conditions that a
user state must meet in order for a function to be called. 
The end of an intervention 
$I$ is always specified by the \texttt{STOP} action.
We also define a cost function, $C: \mathcal{A}\times\mathcal{S} \rightarrow \mathbb{R}$
which mimics the effort made by a given user to perform an action
given the current state. The cost is computed by looking at a causal
graph $\mathcal{G}$, where the nodes of the graph are the user's
features. This assumption encodes the concept of \textit{consequences}
and it ensures a notion of order for the intervention's actions. For
example, it might be easier to get first a degree and then a better
salary, rather than doing the opposite. Fig. \ref{fig:causal_graph} shows an example of a causal graph $\mathcal{G}$ and of the corresponding costs. 
Our goal is to train an agent that, given a user with an unfavourable
outcome, generates counterfactual interventions that overturn it.
Given a black-box classifier $h$, a user $s_0$ for whom the prediction
by $h$ is unfavourable (i.e., $h(s_0) = False$), a causal graph
$\mathcal{G}$ and a set of possible actions $\mathcal{A}$ (implicitly
represented by the functions in $\mathcal{F}$ and their arguments in $\mathcal{X}$), we
want to generate a sequence $I^*$, that, if applied to $s_0$, produces
a new state, $s^* = I(s_0)$, such that $h(s^*) = True$.
This sequence must be \textit{actionable}, which means that
the user has to be able to perform those actions, and minimize the
user's \textit{cost}.  More formally:

\begin{eqnarray}
    & I^* = \min_{I} & \sum_{t=0}^T C(a_t,s_t) \\ 
    & \hbox{s.t.} & I = \{a_t\}_{t=0}^T \quad a_t \in \mathcal{A} \quad \forall t \nonumber \\
    %&&  \nonumber \\
    && s_t = I_{t-1}(s_{t-1}) \quad \forall t > 0 \nonumber \\
    &&  h(I(s_0)) \neq h(s_0) \nonumber
\end{eqnarray}

\begin{figure}[t!]
    \centering
    \includegraphics[width=0.5\linewidth]{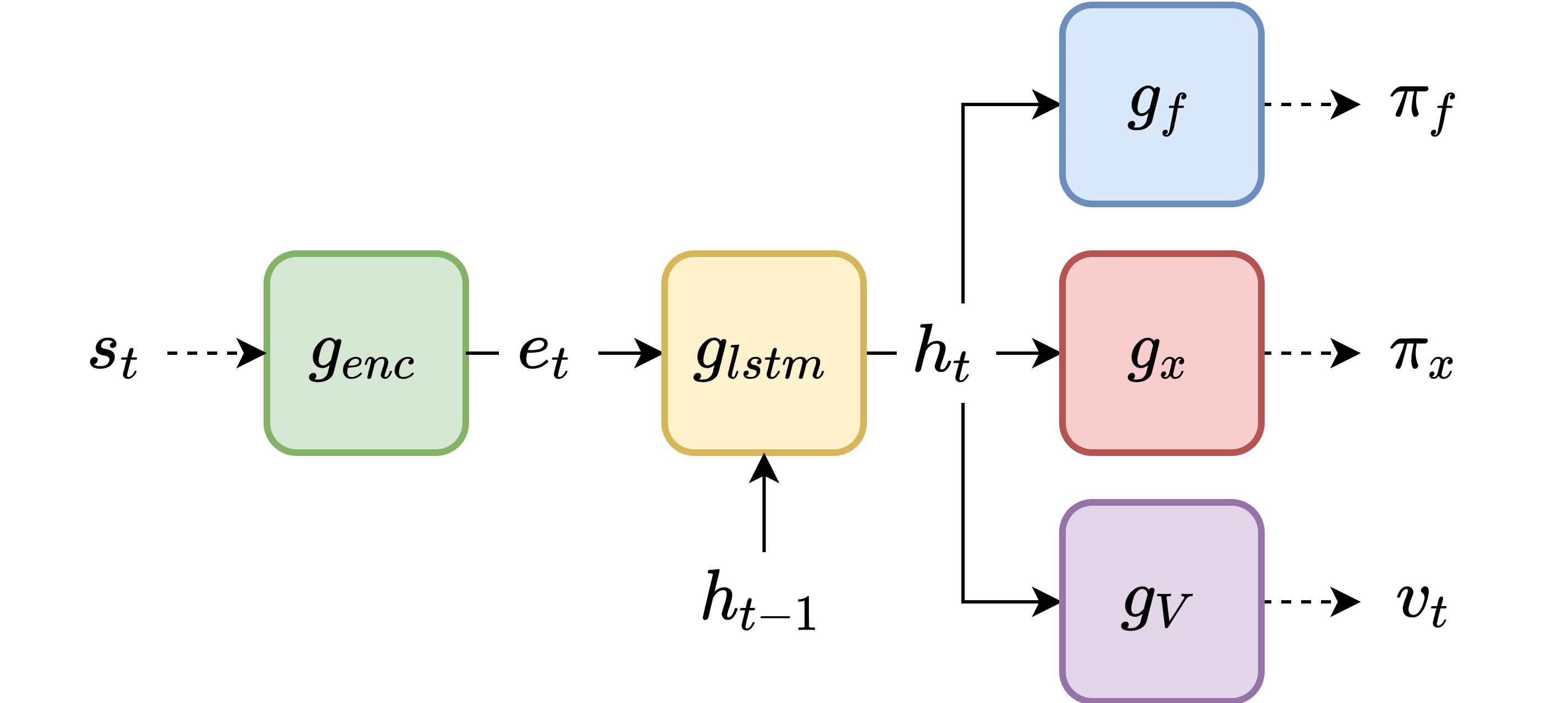}
    \caption{\textbf{Agent architecture}. Given the user's state $s_t$, it outputs a function policy, $\pi_f$, an argument policy $\pi_x$ and an estimate of the expected reward from the state $v_t$. These outputs are used to select the next best action $(f,x)_{t+1}$.}
    \label{fig:agent_architecture}
\end{figure}

\subsection{Model Architecture}

\paragraph{Overall structure.} Fig. \ref{fig:complete_architecture} shows the complete model architecture. It is composed of a binary encoder and an RL agent coupled with the Monte Carlo Tree Search procedure. The binary encoder converts the user's features into a binary representation. The conversion is done by one-hot-encoding the categorical features and discretizing the numerical features into ranges. In the following sections, we will use $s_t$ to directly indicate the user's state binary version. Given a state $s_t$, the RL agent generates candidate policies, $\pi_{f}$ and $\pi_{x}$, for the function and argument generation respectively. MCTS uses these policies as priors for its exploration of the action space and extracts the best next action $(f, x)_{t+1}^*$. The action is then applied to the environment. The procedure ends when the \texttt{STOP} action is chosen (i.e., the intervention was successful) or when the maximum intervention length is reached, in which case the result is marked as a failure. During training, the reward is used to improve the MCTS estimates of the policies. Moreover, correct traces (i.e., traces of interventions leading to the desired outcome change) are stored in a replay buffer, and a sample of traces from the buffer is used to refine the RL agent. 
      
\paragraph{RL agent structure.} The agent structure is inspired by previous program synthesis works \cite{detoni2021learning,pierrot2019learning}. It is composed by 5 components: a state encoder, $g_{enc}$, an LSTM controller, $g_{lstm}$, a function network $g_{f}$, an argument network $g_x$ and a value network $g_{V}$. We use simple feedforward networks to implement $g_f$, $g_x$ and $g_V$.
\begin{align}
    g_{enc}(s_t) = e_t \qquad g_{lstm}(e_t, h_{t-1}) = h_{t} \\
    g_{f}(h_{t}) = \pi_{f} \quad g_{x}(h_{t}) = \pi_{x} \quad g_{V}(h_{t}) = v_t
\end{align}
$g_{enc}$ encodes the user's state in a latent representation which is fed to the controller, $g_{lstm}$. The controller, $g_{lstm}$ learns an implicit representation of the program to generate the interventions. The function and argument networks are then used to extract the corresponding policies, $\pi_f$ and $\pi_x$, by taking as input the hidden state $h_t$ from $g_{lstm}$. $g_V$ represents the value function $V$ and it outputs the expected reward from the state $s_t$. Here, we omit the state $s_t$ when defining the policies and the value function output, since $s_t$ is already embedded into the $h_t$ representation.
In our settings, we try to learn a single program, which we call \texttt{INTERVENE}.

\paragraph{Policy.} A policy is a distribution over the available
actions (i.e., functions and their arguments) such that
$\sum_{i=0}^{N} \pi(i) = 1$. Our agent produces two policies:
$\pi_{f}$ on the function space, and $\pi_{x}$ on the argument space.
The next action, $(f,x)_{t+1}$, is chosen by taking the argmax over
the policies:
\[
    f_{t+1} = \argmax_{f \in \mathcal{F}} \pi_{f}(f) \quad x_{t+1} = \argmax_{x \in \mathcal{X}_{f_{t+1}}} \pi_{x}(x\vert f_{t+1})
\]
Each program starts by calling the program \texttt{INTERVENE}, and it ends when the action \texttt{STOP} is called.

\paragraph{Reward.} Once we have applied the intervention $I$, given the black-box classifier $h$, the reward, $r$, is computed as:
\begin{equation}
    r = \lambda^T R \; \quad  \lambda \in (0,1), \;
    R = \begin{cases}
        1 \quad h(I(s)) \neq h(s)\\
        0 \quad \hbox{otherwise}
    \end{cases}
\end{equation} 
where $\lambda$ is a regularization coefficient and $T$ is the length of the intervention. The $\lambda^T$ penalizes longer interventions in favour of shorter ones.

\subsection{Monte Carlo Tree Search}

Monte Carlo Tree Search (MCTS) is a discrete heuristic search
procedure that can successfully solve combinatorial optimization
problems with large action spaces \cite{silver2018alphazero,silver2016alphago}. MCTS explores the
most promising nodes by expanding the search space based on a random
sampling of the possible actions. In our setting, each tree node
represents the user's state at a time $t$, and each arc represents a
possible action determining a transition to a new state. MCTS
searches for the correct sequence of interventions that minimize the
user effort and changes the prediction of the black-box model. We use
the agent policies, $\pi_f$ and $\pi_x$, as a prior to explore the
program space. Then, the newly found sequence of interventions is
used to train the RL agent. To select the next node, we maximize the
UCT criterion \cite{kocsis2006uct}:
\smallskip
\begin{equation}
    (f,x)_{t+1} = \argmax_{f \in \mathcal{F}, x \in \mathcal{X}_f} Q(s, (f,x)) + U(s, (f,x))+L(s,(f,x))
\end{equation}
\smallskip
Here $Q(s, (f,x))$ returns the expected reward by taking action $(f,x)$.
$U(s, (f,x))$ is a term which trades-off exploration and exploitation, and it is based on how many times we visited node $s$ in the tree. $L(s, (f,x))$ is a scoring term which is defined as follows:
\smallskip
\begin{equation}
L(s, (f,x)) = {e^{-(l_{cost}((f,x),s) + l_{count}(f))}}
\end{equation}
\smallskip
where $l_{cost}=C(a,s) \in \mathbb{R}$ represents the \textit{effort} needed to perform the $a=(f,x) \in \mathcal{A}$ action, and $l_{count} \in \mathbb{R}$ penalizes interventions that call multiple times the same function $f$.
MCTS uses the simulation results to return an improved version of the agent policies $\pi_f^{mcts}$ and $\pi_x^{mcts}$.

From the found intervention, we build an \textit{intervention trace}, 
which is a sequence of tuples that stores, for each time step $t$: the input
state, the output state, the reward, the hidden state of the
controller and the improved policies.  
The traces are stored in the replay buffer, to be used to train the RL agent.

\subsection{Training the agent}

The agent has to learn to replicate the interventions provided by MCTS at each step $t$.
Given the replay buffer, we sample a batch of \textit{intervention traces} and we minimize the cross-entropy $\mathcal{L}$ between the MCTS policies and the agent policies for each time step $t$:
\begin{equation}
    \argmin_{\theta} \sum_{batch} (V-r)^2 -(\pi_{f}^{mcts})^T\log(\pi_{f})  -(\pi_{x}^{mcts})^T\log(\pi_{x}) \label{eq:loss} 
\end{equation}
where $\theta$ represents the agent's parameters and $V$ is the value function evaluation computed by the agent.

\subsection{Generate Interventions through RL}

When training the agent, we learn a general policy that can be used to provide interventions for many different users. The inference procedure is similar to the one used for training.
Given an initial state $s$, MCTS explores the tree search space using as ``prior" the learnt policies $\pi_x$ and $\pi_f$ coming from the agent. The policies $\pi_x$ and $\pi_f$ give MCTS a hint of which node to select at each step. Once MCTS finds the minimal cost trace that achieves recourse, we return it to the user.
In principle, we can also use only $\pi_x$ and $\pi_f$ to obtain a viable intervention (e.g., by deterministically taking the action with highest probability each time). However, keeping the search component (MCTS) with a small exploration budget outperforms the RL agent alone. See Table \ref{tab:accuracy-agent} in Section 4 for the comparison between the agent-only model and the agent augmented with MCTS.

Learning a general policy to provide interventions is a powerful feature. However, the policy is encoded in the latent states of the agent, thus making it impossible for us to understand it. We want to be able to extract from the trained model an explainable version of this policy, which can then be used to \textit{explain} why the model suggested a given intervention. Namely, besides providing to the users a sequence of actions, we want to show also the reason behind each suggested action. The intuition to achieve this is the following:  given a set of successful interventions generated by the agent, we can distill a synthetic automaton, or \textit{program}, which condense the policy in a graph-like structure which we can traverse.

\subsection{Explainable Intervention Program}

\begin{figure*}
    \centering
    \includegraphics[width=\textwidth]{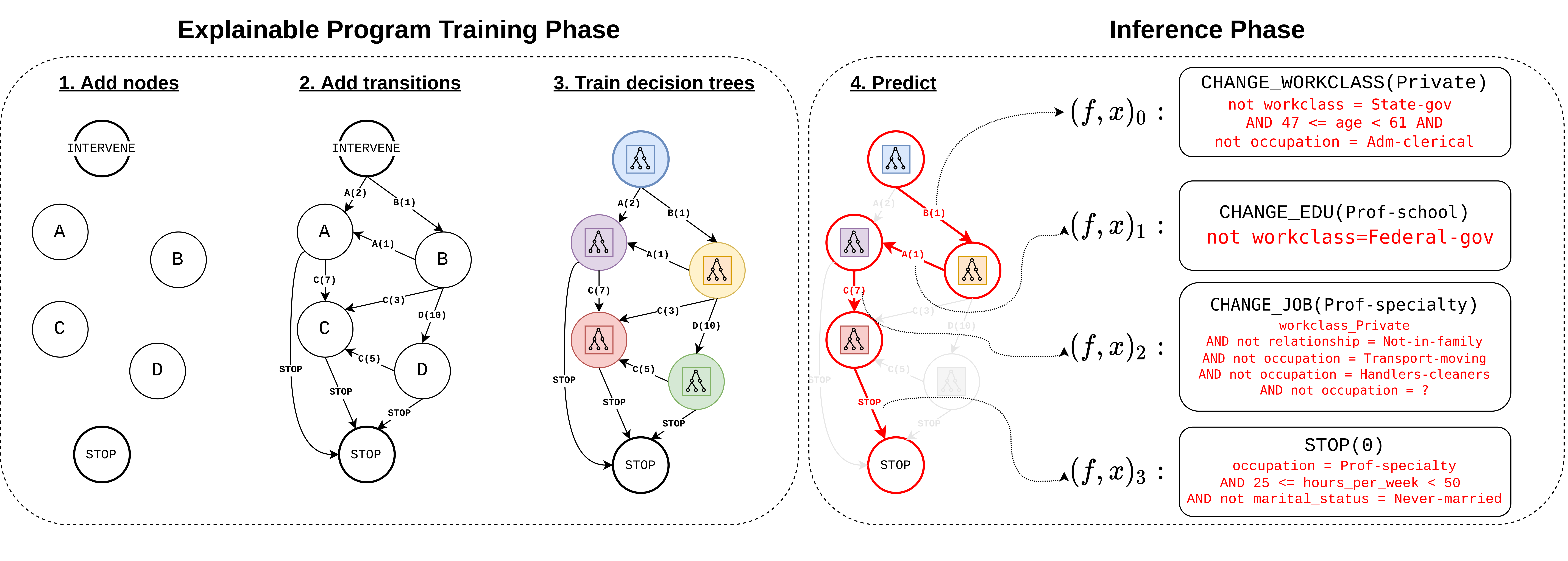}
    \caption{\textbf{Procedure to generate the explainable program from intervention traces.} 1. For all $f \in \mathcal{F}$, we add a new node. 2. Given the samples traces, we add the transitions, and we store $(s_i, (f_i, x_i))$ in each node. 3. We train a decision tree for each node to predict the next action (consistently with the sampled traces). 4. We execute the program on the new instance at prediction time, using the decision trees to decide the next action at each node. We extract a Boolean rule explaining it from the corresponding decision tree for each action. On the right, an example of generated intervention. The actions $(f,x)$ are black, while the explanations are red.}
    \label{fig:exp-program-procedure}
\end{figure*}

We now show how we can build a deterministic program given the agent. Fig. \ref{fig:exp-program-procedure} shows the complete procedure and an example of the produced trace. 
First, we sample $M$ intervention traces from the trained agent and extract a sequence of $\{(s_i, (f, x)_i)\}_{i=0}^T$ for each trace.
Then, we construct an automaton graph, $\mathcal{P}$, in the following way:
\begin{enumerate}
    \item Given the function library $\mathcal{F}$, we create a node for each function $f$ available. We also add a starting node called \texttt{INTERVENE} and a ``sink" node called \texttt{STOP};
    \item We connect each node by unrolling the sampled traces. Starting from \texttt{INTERVENE}, we treat each action $(f,x)_{t}$ as a transition. We label the transition with $(f,x)$ and we connect the current node to the one representing the function $f$;
    \item Lastly, for each node $f$, we store a collection of outgoing state-action pairs $(s_i, (f,x)_i)$. Namely, we store all the states $s$ and the corresponding outward transitions which were decided by the model while at the node $f$;
    \item For each node, $f \in \mathcal{P}$, we train a decision tree on the tuples $(s_i,(f,x)_i)$ stored in the node to predict the transition $(f,x)_i$ given a user's state $s_i$.
\end{enumerate}
The decision trees are trained only once by using the collection of traces sampled from the trained agent. The agent is frozen at this step, and it is not trained further. At this point, we perform Step 1 to 3 of Fig. \ref{fig:exp-program-procedure}.
The pseudocode of the entire procedure is available in the Appendix \ref{app:pseudocode}.

\subsection{Generate Explainable Interventions} 
The intervention generation is done by traversing the graph $\mathcal{P}$, starting from the node \texttt{INTERVENE}, until we reach the \texttt{STOP} node or we reach the maximum intervention length. In the last case, the program is marked as a failure. Given the node $f \in \mathcal{P}$ and given the state $s_t$, we use the decision tree of that node to predict the next transition $(f',x')$. Moreover, we can extract from the decision tree interpretable rules which tell us why the next action was chosen. A \textit{rule} is a boolean proposition on the user's features such as $(income > 5000 \wedge education = bachelor)$. Then, we follow $(f',x')$, which is an arc going from $f$ to the next node $f'$, and we apply the action to $s_t$ to get $s_{t+1}$. Again, the program is ``fixed" at inference time, and it is not trained further. See Step 4 of Fig. \ref{fig:exp-program-procedure} for an example of the inference procedure and of the produced explainable trace.

\section{Experiments}
Our experimental evaluation aims at answering the following research questions:
 (1) Does our method provide better performances than the competitors in terms of the accuracy of the algorithmic recourse? 
 (2) Does our approach allow us to complement interventions with action-by-action explanations in most cases?
 (3) Does our method minimize the interaction with the black-box classifier to provide interventions? 

The code and the dataset of the experiments are available on Github to ensure reproducibility\footnote{\url{https://github.com/unitn-sml/syn-interventions-algorithmic-recourse}}. The software exploit parallelization through \texttt{mpi4python} \cite{dalcin2021mpi4python} to improve inference and training time. We compared the performance of our algorithm with CSCF \cite{naumann_consequence-aware_2021}, to the best of our knowledge the only existing model-agnostic approach that can generate consequence-aware interventions following a causal graph. However, note that earlier solutions still perform user-specific optimization, so that our results in terms of generality, interpretability and cost (number of queries to the black-box classifier and computational cost) carry over to these alternatives. For the sake of a fair comparison, we built our own parallelized version of the CSCF model based on the original code. We developed the project to make it easily extendable and reusable by the research community.
The experiments were performed using a Linux distribution on an Intel(R) Xeon(R) CPU E5-2660 2.20GHz with 8 cores and 100 GB of RAM (only 4 cores were used).

\begin{table}[t]
\centering
\resizebox{0.7\linewidth}{!}{%
\begin{tabular}{lrrrrrr}
\toprule
\textbf{Dataset} & $\vert D\vert$ & $h(s)=1$ & $h(s)=0$ & $\vert s\vert$ & $ \vert B(s) \vert$ & $ \vert\mathcal{F} \vert$ \\ \midrule
\textit{german} & 1002 & 301 & 701 & 10 & 44 & 7 \\
\textit{adult} & 48845 & 11691 & 37154 & 15 & 125 & 6 \\
\textit{syn} & 10004 & 5002 & 5002 & 10 & 40 & 6 \\
\textit{syn\_long} & 10004 & 5002 & 5002 & 14 & 64 & 10 \\
\bottomrule
\end{tabular}%
}
\caption{\textbf{Description of the datasets.} $\vert D \vert$ is the size of the dataset. $\vert s\vert$ the number of features for an instance. $\vert B(s)\vert$ shows how many binary features the agent sees after the conversion with the binary converter. $\vert \mathcal{F}\vert$ is the size of the agent program library. $h(s)$ indicates the number of favourable (1) and unfavourable (0) samples.}
\label{tab:dataset}
\end{table}

\begin{figure*}[t!]
    \centering
    \includegraphics[width=\linewidth]{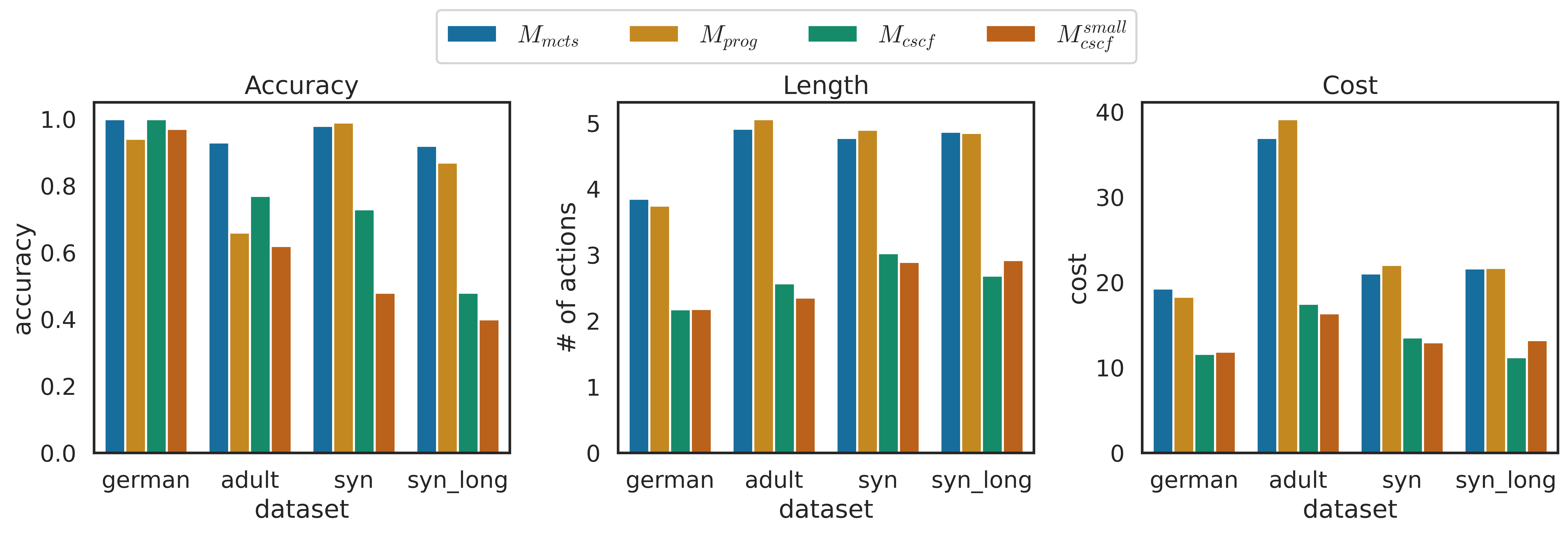}
    \caption{\textbf{Experimental results.} (Left) Accuracy (fraction of successful interventions); (Middle) Average length of a successful intervention; (Right) Average cost of a successful intervention. Results are averaged over 100 test examples.}
    \label{fig:results}
\end{figure*}

\subsection{Dataset and black-box classifiers}

Table \ref{tab:dataset} shows a brief description of the datasets. They all represent binary (favourable/unfavourable) classification problems. The two real world datasets, \textit{German Credit} (\textit{german}) and \textit{Adult Score} (\textit{adult}) \cite{Dua:2019}, are taken from the relevant literature.
Given that in these datasets a couple of actions is usually sufficient to overturn the outcome of the black-box classifier, we also developed two synthetic datasets, \textit{syn} and \textit{syn\_long}, where longer interventions are required, so as to evaluate the models in more challenging scenarios.
The datasets are made by both categorical and numerical features (e.g., monthly income, job type, etc.). Each dataset was randomly split into $80\%$ train and $20\%$ test.
For each dataset, we manually define a causal graph, $\mathcal{G}$, by looking at the features available. For the synthetic datasets, we sampled instances directly from the causal graph. 
The black-box classifier for \textit{german} and \textit{adult} was obtained by training a 5-layers MLP with ReLu activations.
The trained classifiers are reasonably accurate ($\sim0.9$ test-set accuracy for \textit{german}, $\sim0.8$ for \textit{adult}).
The synthetic datasets (\textit{syn} and \textit{syn\_long}) do not require any training since we directly use our manually defined decision function.

\subsection{Models}

We evaluate four different models: the agent coupled with MCTS ($M_{mcts}$), the explainable deterministic program distilled from the agent ($M_{prog}$), and two versions of CSCF, one ($M_{cscf}$) with a large budget of generation, $n$, and population size, $p$, ($n=50, p=200$) and one ($M_{cscf}^{small}$) with a smaller budget ($n=25, p=100$). 

\subsection{Evaluation}

The left plot in Fig. \ref{fig:results} shows the average accuracy of the different models, namely the fraction of instances for which a model manages to generate a successful intervention. 
We can see how $M_{mcts}$ outperforms or is on-par with the $M_{cscf}$ and $M_{cscf}^{small}$ models on both the real-word and synthetic datasets. The performance difference is more evident in the synthetic datasets, because the evolutionary algorithm struggles in generating interventions that require more than a couple of actions. The accuracy loss incurred in distilling $M_{mcts}$ into a program ($M_{prog}$) is rather limited. This implies that we are able to provide interventions with explanations for $94\%$ (\textit{german}), $66\%$ (\textit{adult}), $99\%$ (\textit{syn}) and $87\%$ (\textit{syn\_long}) of the test users\footnote{Note that the accuracy loss observed on \textit{adult} is due to the limited sampling budget we allocated for $M_{prog}$ ($250$ traces for all datasets). Adapting this budget to the feature space size (considerably larger for \textit{adult}) can help boosting the performance, at the cost of generating longer explanations.}. Moreover, $M_{prog}$ generates similar interventions to $M_{mcts}$. The sequence similarity between their respective interventions for the same user are $0.89$ (\textit{german}), $0.72$ (\textit{adult}), $0.80$ (\textit{syn}) and $0.71$ (\textit{syn\_long}), where $1.0$ indicates identical interventions.

\begin{figure*}
    \centering
    \includegraphics[width=0.8\linewidth]{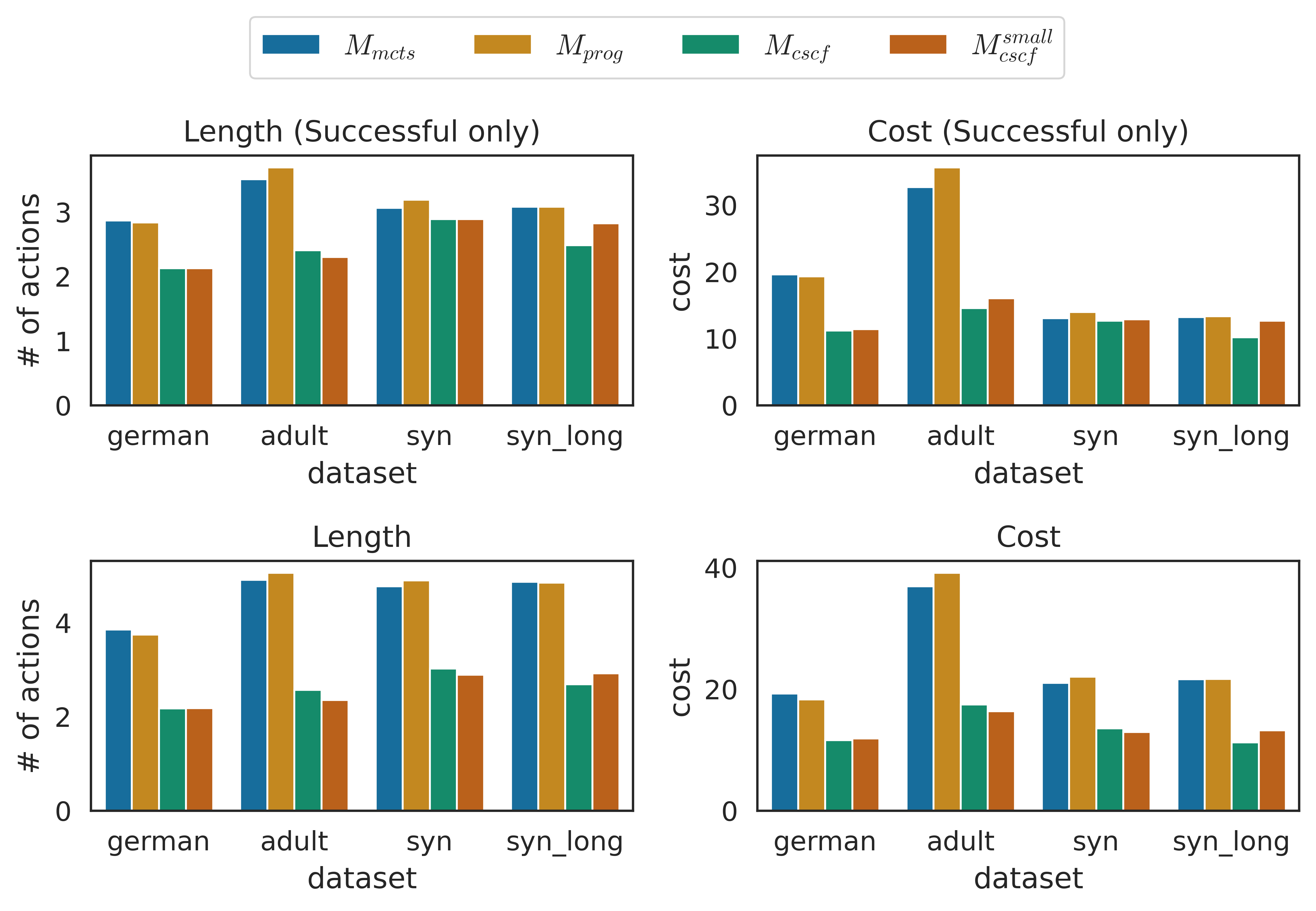}
    \caption{\textbf{Evaluation considering only the instances for which all the models provide a successful intervention.} If we restrict the comparison to the subset of instances for which all models manage to generate a successful intervention, the difference in costs between methods shrinks substantially (top left vs bottom left). The same behaviour applies to the intervention length (top right vs bottom right).}
    \label{fig:accuracy_intersection}
\end{figure*}

The main reason for the accuracy gains of our model is the ability to generate long interventions, something evolutionary-based algorithms struggle with.
This effect can be clearly seen from the middle plot of Fig.~\ref{fig:results}. Both $M_{cscf}$ and $M_{cscf}^{small}$ rarely generate interventions with more than two actions, while our approach can easily generate interventions with up to five actions. A drawback of this ability is that intervention costs are, on average, higher (right plot of Fig.~\ref{fig:results}). On the one hand, this is due to the fact that our model is capable of finding interventions for more complex instances, while $M_{cscf}$ and $M_{cscf}^{small}$ fail. Indeed, if we compute lengths and costs on the subset of instances for which all models find a successful intervention, the difference between the approaches is less pronounced. See Fig. \ref{fig:accuracy_intersection} for the evaluation. On the other hand, there is a clear trade-off between solving a new optimization problem from scratch for each new user, and learning a general model that, once trained, can generate interventions for new users in real-time and without accessing the black-box classifier.

\begin{table*}[]
\centering
\resizebox{0.3\textwidth}{!}{%
\begin{tabular}{lcc}
\toprule
\textbf{Dataset} & \textbf{$M_{mcts}$} & \textbf{$M_{agent}$} \\
\midrule
\textit{german} & \textbf{1.00} & 0.00 \\
\textit{adult} & \textbf{0.93} & 0.00 \\
\textit{syn} & \textbf{0.98} & 0.59 \\
\textit{syn\_long} & \textbf{0.92} & 0.00\\
\bottomrule
\end{tabular}%
}
\caption{\textbf{Ablation study}.  
In order to evaluate the contribution of MCTS in finding successful interventions, we also developed an additional model which only uses the RL agent, $M_{agent}$. The agent just predicts the next action using its own policy without leveraging MCTS to refine the choice. Results indicate that RL alone is incapable of finding successful interventions.}
\label{tab:accuracy-agent}
\end{table*}

\begin{figure}[]
    \centering
    \includegraphics[width=0.6\linewidth]{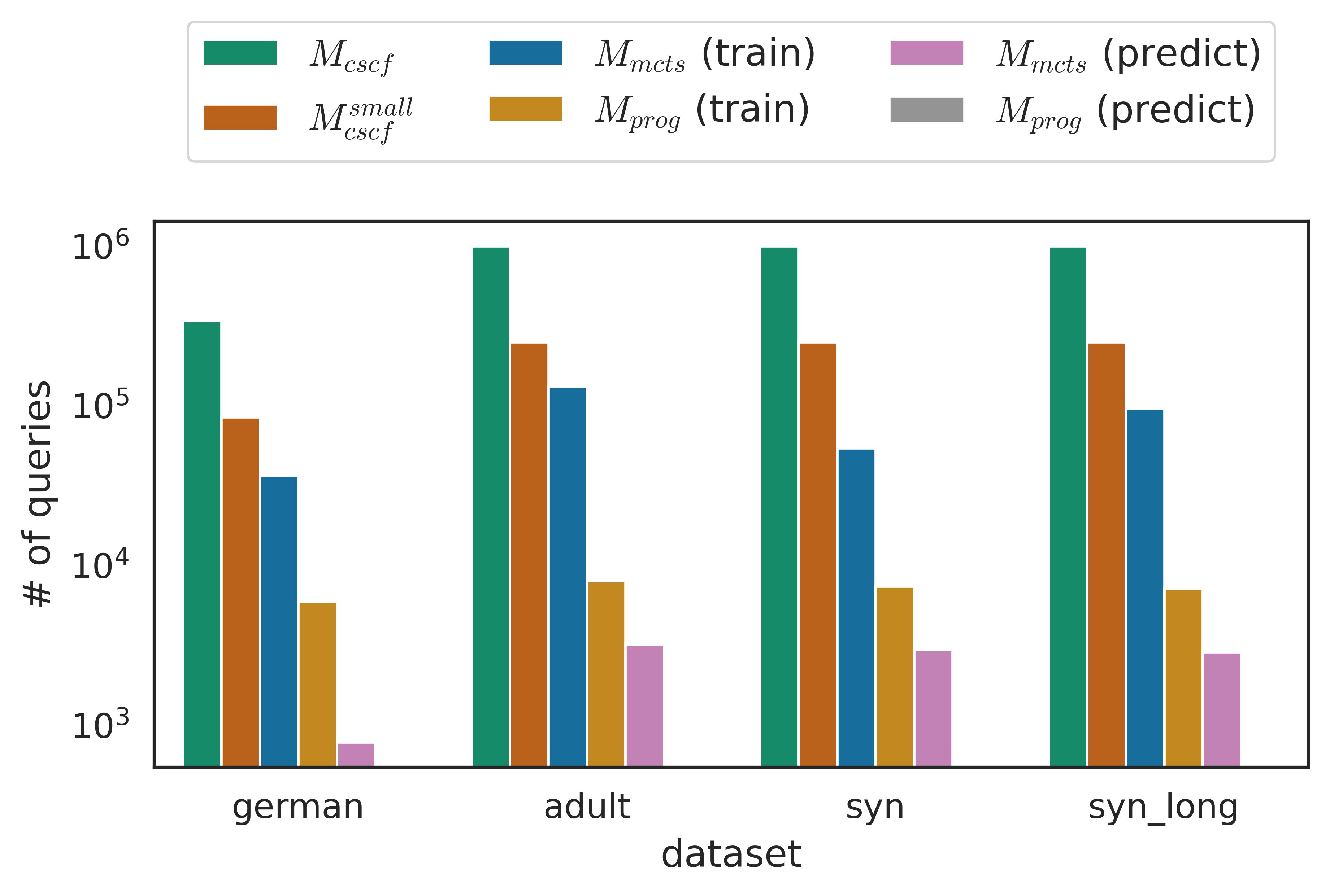}
    \caption{\textbf{Number of queries.} Total number of queries to the black-box classifier made by the models. $M_{prog}(predict)$ is not visible, as the automaton does not query the black-box classifier to generate interventions. Note that the number of queries is in logscale.}
    \label{fig:blackbox-calls}
\end{figure}

\begin{figure*}
    \centering
    \includegraphics[width=\linewidth]{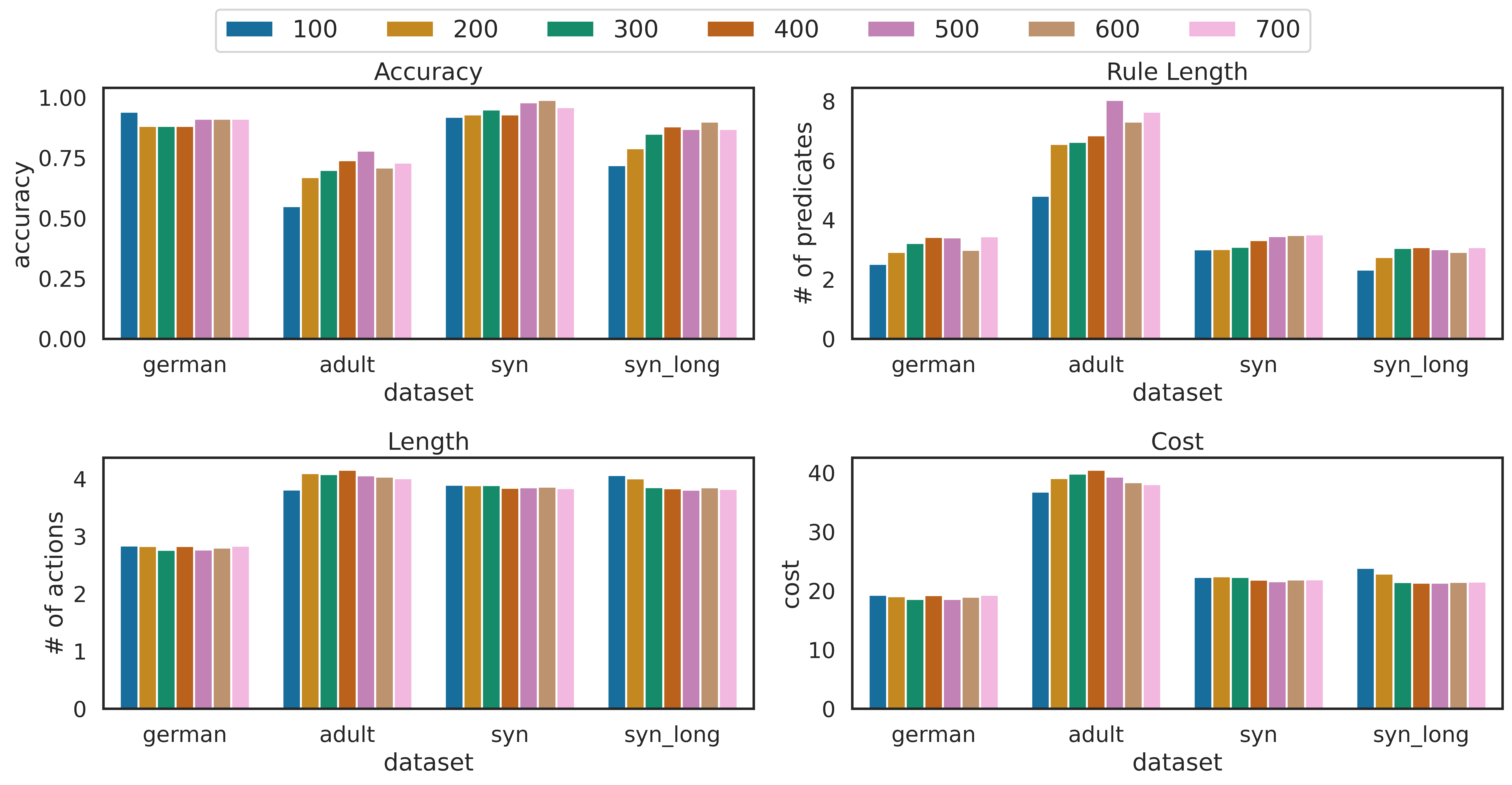}
    \caption{\textbf{Accuracy of $M_{prog}$ when varying the training budget.} We show the effect on increasing the sampling budget (from 100 to 700 traces) when training the $M_{prog}$ model.}
    \label{fig:accuracy_program}
\end{figure*}

Fig. \ref{fig:blackbox-calls} reports the average number of queries to the black-box classifier. Our approach requires far fewer queries than $M_{cscf}$ (note that the plot is in logscale), and even substantially less than $M_{cscf}^{small}$ (that is anyhow not competitive in terms of accuracy). Furthermore, most queries are made for training the agent ($M_{mcts}(train)$), which is only done once for all users. Once the model is trained, generating interventions for a single user requires around two orders of magnitude fewer queries than the competitors. Note that MCTS is crucial to allow the RL agent to learn a successful policy with a low budget of queries. Indeed, training an RL agent without the support of MCTS fails to converge in the given budget, leading to a completely useless policy. By efficiently searching the space of interventions, MCTS manages to quickly correct inaccurate initial policies, allowing the agent to learn high quality policies with a limited query budget.  See Table \ref{tab:accuracy-agent} for the evaluation.

When turning to the program, building the automaton ($M_{prog}(train)$) requires a negligible number of queries to extract the intervention traces used as supervision.

Using the automaton to generate interventions {\em does not} require to query the black-box classifier. This characteristic can substantially increase the usability of the system, as $M_{prog}$ can be employed directly by the user even if they have no access to the classifier. Computationally speaking, the advantage of a two-step phase is also quite dramatic. $M_{cscf}$ takes an average of $\sim 693 s$ for each user to provide a solution (the same order of magnitude of training a model for all users with $M_{mcts}$), while $M_{mcts}$ inference time is under $1 s$, allowing real-time interaction with the user.

Additionally, Fig. \ref{fig:accuracy_program} shows how it is possible to improve the performances of $M_{prog}$ by just sampling more traces from the trained agent ($M_{mcts}$).
We can see how the accuracy increases in the \textit{adult}, \textit{syn} and \textit{syn\_long} datasets. We also notice that using a larger budget to train $M_{prog}$ produces longer explainable rules by keeping the length and cost of the generated interventions almost constant. The total number of queries to the black-box classifier will also slightly increase.

Overall, our experimental evaluation allows us to affirmatively answer the research questions stated above.

\section{Conclusion}
This work improves the state-of-the-art on algorithmic recourse by providing a method that can generate effective and interpretable counterfactual interventions in real-time. Our experimental evaluation confirms the advantages of our solution with respect to alternative consequence-aware approaches in terms of accuracy, interpretability and number of queries to the black-box classifier.
Our work unlocks many new research directions, which could be explored to solve some of its limitations. First, following previous work on causal-aware intervention generation, we use manually-crafted causal graphs and action costs. Learning them from the available data directly, minimizing the human intervention, would allow applying the approach in settings where this information is not available or unreliable.
Second, we showed how our method learns a general program by optimizing over multiple users. It would be interesting to investigate additional RL methods to optimize the interventions globally and locally to provide more personalized sequences to the users. Such methods could be coupled with interactive approaches eliciting preferences and constraints directly from the user, thus maximizing the chance to generate the most appropriate intervention for a given user.

\section*{Ethical Impact}

The research field of algorithmic recourse aims at improving fairness, by providing unfairly treated users with tools to overturn unfavourable outcomes. By providing real-time, explainable interventions, our work makes a step further in making these tools widely accessible. As for other approaches providing counterfactual interventions, our model could in principle be adapted by malicious users to ``hack" a fair system. Research on adversarial training can help in mitigating this risk.

\backmatter

\bmhead{Acknowledgments}
This research was partially supported by TAILOR, a project funded by EU Horizon 2020 research and innovation programme under GA No 952215.

\section*{Declarations}

\bmhead{Funding}
This research was partially supported by TAILOR, a project funded by EU Horizon 2020 research and innovation programme under GA No 952215.

\bmhead{Conflict of interest/Competing interests}
The authors have no competing interests to declare that are relevant to the content of this article.

\bmhead{Ethics approval} 
Not applicable

\bmhead{Consent to participate}
Not applicable

\bmhead{Consent for publication}
Not applicable

\bmhead{Availability of data and materials}
The datasets used in the experimental evaluation are freely available at \url{https://github.com/unitn-sml/syn-interventions-algorithmic-recourse}.

\bmhead{Code availability} 
The code is freely available at \url{https://github.com/unitn-sml/syn-interventions-algorithmic-recourse}.

\bmhead{Authors' contributions} GDT designed the method, conducted the data collection process, built the experimental infrastructure and performed the relevant experiments. BL and AP contributed to the design of the method, provided supervision and resources. All authors contributed to the writing of the manuscript.

\noindent
%If any of the sections are not relevant to your manuscript, please include the heading and write `Not applicable' for that section. 

%%===================================================%%
%% For presentation purpose, we have included        %%
%% \bigskip command. please ignore this.             %%
%%===================================================%%

\begin{appendices}

\section{Program Distillation Pseudocode}\label{app:pseudocode}

We present here the pseudocode of two algorithms. Algorithm \ref{alg:algorithm-generate-program} shows how to distill the synthetic program from the agent and it refers to Step 3 of Fig. \ref{fig:exp-program-procedure}. Algorithm \ref{alg:algorithm-generate-program} shows how the distilled program is applied at inference time to a new user and it refers to Step 4 of Fig. \ref{fig:exp-program-procedure}.

\begin{algorithm*}[]
\caption{\textbf{Generate the explainable program}. Given the automaton reconstructed from the traces $\hat{\mathcal{P}}$, we generate the explainable program $\mathcal{P}$. For each node, we train the decision tree classifier on the traces stored $(s, (f,x))_i$ to predict $(f,x)$ from $s$.}
\label{alg:algorithm-generate-program}
\textbf{Input}: $\hat{\mathcal{P}}$, automaton extracted from the traces\\
\textbf{Output}: $\mathcal{P},$ explainable program
\begin{algorithmic}[1] %[1] enables line numbers
\State Let $t=0$.
\State Let $\mathcal{P} = \{\}$
\ForAll{$node \in \hat{\mathcal{P}}$}
\State{$program = node.program$}
\If {$program != STOP$}
    \If {\textit{node.arcs.unique()} $>$ 1}
        \State \textit{Train a decision tree, $\mathcal{D}$, on node.arcs}
        \State{$node.classifier = \mathcal{D}$}
    \Else
        \State{$node.classifier = \{True \rightarrow node.arcs[0] \}$}
    \EndIf
    \State{$\mathcal{P}.append(node)$}
\EndIf
\EndFor
\State \textbf{return} $\mathcal{P}$
\end{algorithmic}
\end{algorithm*}

\begin{algorithm*}[]
\caption{\textbf{Predict with the program}. Given the program $\mathcal{P}$, we can infer the complete explainable intervention for a user $s_0$ by traversing $\mathcal{P}$. This can be done recursively. The recursion ends when we either predict the \texttt{STOP} action or we reach the maximum intervention length, $\alpha \in \mathbb{N}^+$.}
\label{alg:run-program}
\textbf{Input}: $\mathcal{P}$, $s_0$, $(f,x)_0$, $\mathcal{I}$, $\mathcal{R}$ \\
\textbf{Output}: $s$,the final state, $\mathcal{I}$,the intervention actions, $\mathcal{R}$, rules for each action
\begin{algorithmic}[1] %[1] enables line numbers
\If{$(f,x)_t == (STOP, 0) \; \vee \; t > \alpha $}
 \State \textbf{return} $s_{t+1}, \mathcal{I}, \mathcal{R}$
\Else
    \State $(f,x)_{t+1}, \; \; rule_{t+1} = \mathcal{P}.get(f).predict(s_t)$
    \State $s_{t+1} = (f,x)_{t}.apply(s_t)$
    \State 
    \State $\mathcal{R}.append(rule_{t+1})$
    \State $\mathcal{I}.append((f,x)_{t+1})$
    \State \textbf{recursive call} ($\mathcal{P}, s_{t+1} (f,x)_{t+1}, \mathcal{I}, \mathcal{R}$)
\EndIf
\end{algorithmic}
\end{algorithm*}

\section{Domain Specific Languages (DSL)}\label{app:actions}

We now show some Domain Specific Languages (DSL) used for the \textit{german} (Table \ref{tab:dsl-german}) and \textit{synthetic} experiments. For each setting, we show the functions available, the argument type they accept and an exhaustive list of the potential arguments. Each program operates on a single feature. The name of the program suggests the feature it operates on (e.g., \textit{CHANGE\_JOB} operate on the feature \textit{job}). The programs which accept numerical arguments simply add their argument to the current value of the target feature. The program \textit{STOP} does not accept any argument and it signals only the end of the intervention without changing the features. The DSLs for the \textit{adult} and \textit{synthetic\_long} experiments are similarly defined and are omitted for brevity.

\begin{table}[h!]
\centering
\resizebox{\textwidth}{!}{%
\begin{tabular}{ccp{7cm}}
\toprule
\textbf{Program} & \textbf{Argument Type} & \textbf{Argument}  \\
\midrule
CHANGE\_SAVINGS & Categorical & unknown, little,  moderate, rich, quite\_rich \\
CHANGE\_JOB & Categorical & unskilled\_non\_resident, unskilled\_resident, skilled, highly\_skilled \\
CHANGE\_CREDIT & Numerical & 100, 1000, 2000, 5000 \\
CHANGE\_HOUSING & Categorical & free, rent, own \\
CHANGE\_DURATION & Numerical & 10, 20, 30 \\
CHANGE\_PURPOSE & Categorical & business, car, domestic\_appliances, education, furniture/equipment, radio/TV, repairs, vacation/others \\
STOP & - & - \\  \bottomrule
\end{tabular}%
}
\caption{Example of the DSL for the \textit{german} experiment.}
\label{tab:dsl-german}
\end{table}

\begin{table}[h!]
\centering
\resizebox{\textwidth}{!}{%
\begin{tabular}{ccc}
\toprule
\textbf{Program} & \textbf{Argument Type} & \textbf{Argument} \\ \midrule
CHANGE\_EDUCATION & Categorical & none,secondary school diploma, bachelor, master, phd \\
CHANGE\_JOB & Categorical & unemployed, worker, office worker, manager, ceo \\
CHANGE\_INCOME & Numerical & 5000, 10000, 20000, 30000, 40000, 50000 \\
CHANGE\_HOUSE & Categorical & none, rent, own \\
CHANGE\_RELATION & Categorical & single, married, divorced, widow/er \\
STOP & - & -\\
\bottomrule
\end{tabular}%
}
\caption{Example of the DSL for the \textit{synthetic} experiment.}
\label{tab:dsl-synthetic}
\end{table}

\end{appendices}

%%=============================================%%
%% For submissions to Nature Portfolio Journals %%
%% please use the heading ``Extended Data''.   %%
%%=============================================%%

%%=============================================================%%
%% Sample for another appendix section			       %%
%%=============================================================%%

%% \section{Example of another appendix section}\label{secA2}%
%% Appendices may be used for helpful, supporting or essential material that would otherwise 
%% clutter, break up or be distracting to the text. Appendices can consist of sections, figures, 
%% tables and equations etc.

%%===========================================================================================%%
%% If you are submitting to one of the Nature Portfolio journals, using the eJP submission   %%
%% system, please include the references within the manuscript file itself. You may do this  %%
%% by copying the reference list from your .bbl file, paste it into the main manuscript .tex %%
%% file, and delete the associated \verb+\bibliography+ commands.                            %%
%%===========================================================================================%%

% common bib file
\bibliography{ijclr22}
%% if required, the content of .bbl file can be included here once bbl is generated
%%\input sn-article.bbl

%% Default %%
%%\input sn-sample-bib.tex%

\end{document}